\documentclass{article}

\usepackage[normalem]{ulem}
\useunder{\uline}{\ul}{}

\usepackage{caption}

\usepackage[shortlabels]{enumitem} 

\usepackage{amssymb}
\usepackage{pifont}
\newcommand{\cmark}{\ding{51}}%
\newcommand{\xmark}{\ding{55}}%

\PassOptionsToPackage{numbers,sort,compress}{natbib}

\usepackage[final]{neurips_2025}

\usepackage{hyperref}
\usepackage{url}
\usepackage{booktabs}
\usepackage{amsfonts}
\usepackage{nicefrac} 
\usepackage{microtype} 
\usepackage{xcolor} 
\usepackage{graphicx} 
\usepackage{amsmath}
\usepackage[table,xcdraw]{xcolor}

\title{Towards Automated Petrography}

\author{
  Isai Daniel Chac\'{o}n\thanks{These authors contributed equally to this work.} \\
  Universidad de los Andes, Colombia \\
  \And
  Paola Ruiz Puentes\footnotemark[1] \\
  Universidad de los Andes, Colombia \\
  \And
  Jillian Pearse \\
  California State University, Long Beach \\
  \And
  Pablo Arbel\'{a}ez \\
  Universidad de los Andes, Colombia
}

\makeatletter
\providecommand{\@trackname}{Neurips 2025 Datasets and Benchmarks Track}%
\makeatother

\begin{document}

\maketitle

\begin{abstract}
Petrography is a branch of geology that analyzes the mineralogical composition of rocks from microscopical thin section samples. It is essential for understanding rock properties across geology, archaeology, engineering, mineral exploration, and the oil industry. However, petrography is a labor-intensive task requiring experts to conduct detailed visual examinations of thin section samples through optical polarization microscopes, thus hampering scalability and highlighting the need for automated techniques. To address this challenge, we introduce the Large-scale Imaging and Thin section Optical-polarization Set (LITHOS), the largest and most diverse publicly available experimental framework for automated petrography. LITHOS includes 211,604 high-resolution RGB patches of polarized light and 105,802 expert-annotated grains across 25 mineral categories. Each annotation consists of the mineral class, spatial coordinates, and expert-defined major and minor axes represented as intersecting vector paths, capturing grain geometry and orientation. We evaluate multiple deep learning techniques for mineral classification in LITHOS and propose a dual-encoder transformer architecture that integrates both polarization modalities as a strong baseline for future reference. Our method consistently outperforms single-polarization models, demonstrating the value of polarization synergy in mineral classification. We have made the LITHOS Benchmark publicly available, comprising our dataset, code, and pretrained models, to foster reproducibility and further research in automated petrographic analysis.

\includegraphics[height=1em]{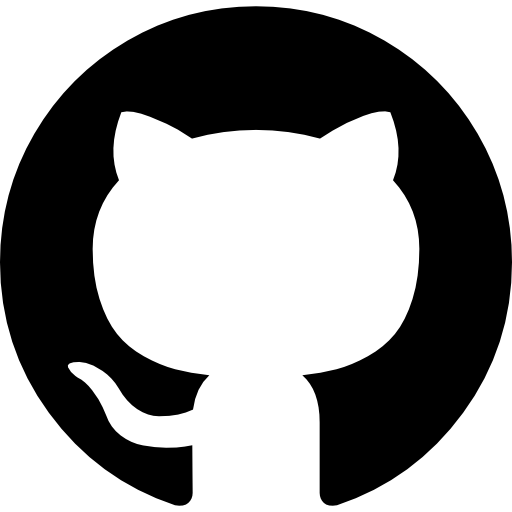}\hspace{0.3em}%
\textbf{Benchmark and code:} \href{https://github.com/BCV-Uniandes/LITHOS}{https://github.com/BCV-Uniandes/LITHOS}

\includegraphics[height=1em]{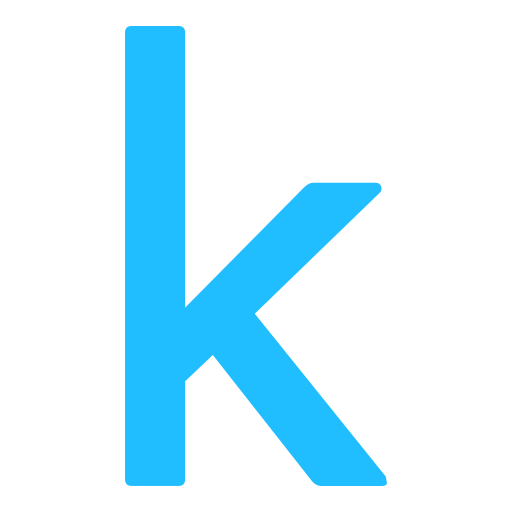}\hspace{0.3em}%
\textbf{Dataset:}\href{https://kaggle.com/datasets/f1caa92f0ca6bace86fc8fa6d557d3cac4d2cfe545f5aeb7ea2f141ab68d3720}{LITHOS Dataset}

\end{abstract}

\section{Introduction}

Petrography is a branch of geology that describes and analyzes rocks and soils, focusing on their mineralogical composition, texture, and structure. It enables the identification of minerals and other constituents within a sample and the characterization of grain size, shape, and porosity distribution \citep{Petrographydef}. These attributes are critical in various fields, particularly in economic geology, where petrographic analysis informs resource exploration and extraction strategies \citep{EconomicGeology,EconomicGeology_exa}.

Understanding a rock's petrographic properties is fundamental across multiple disciplines. In geology, petrography aids in reconstructing Earth's geological history, revealing past environmental conditions and tectonic processes \citep{geology,geology2}. In archaeology, it helps determine the provenance of artifacts and ancient trade routes by analyzing ceramic and lithic materials \citep{archaeology}. In engineering, petrographic examination is crucial for assessing the durability and quality of construction materials, such as concrete and aggregates\citep{concrete,concrete2}. In mineral exploration, it provides essential insights into the characterization of the ore, directly influencing the beneficiation and metallurgical processes \citep{Ore,geology}. Additionally, in the oil industry, petrography plays a key role in evaluating a rock's capacity to store and transmit hydrocarbons, guiding reservoir characterization and exploration efforts \citep{geology,Hydrocarbon}.

\begin{figure}[t]
    \centering
    \includegraphics[width=0.80\linewidth]{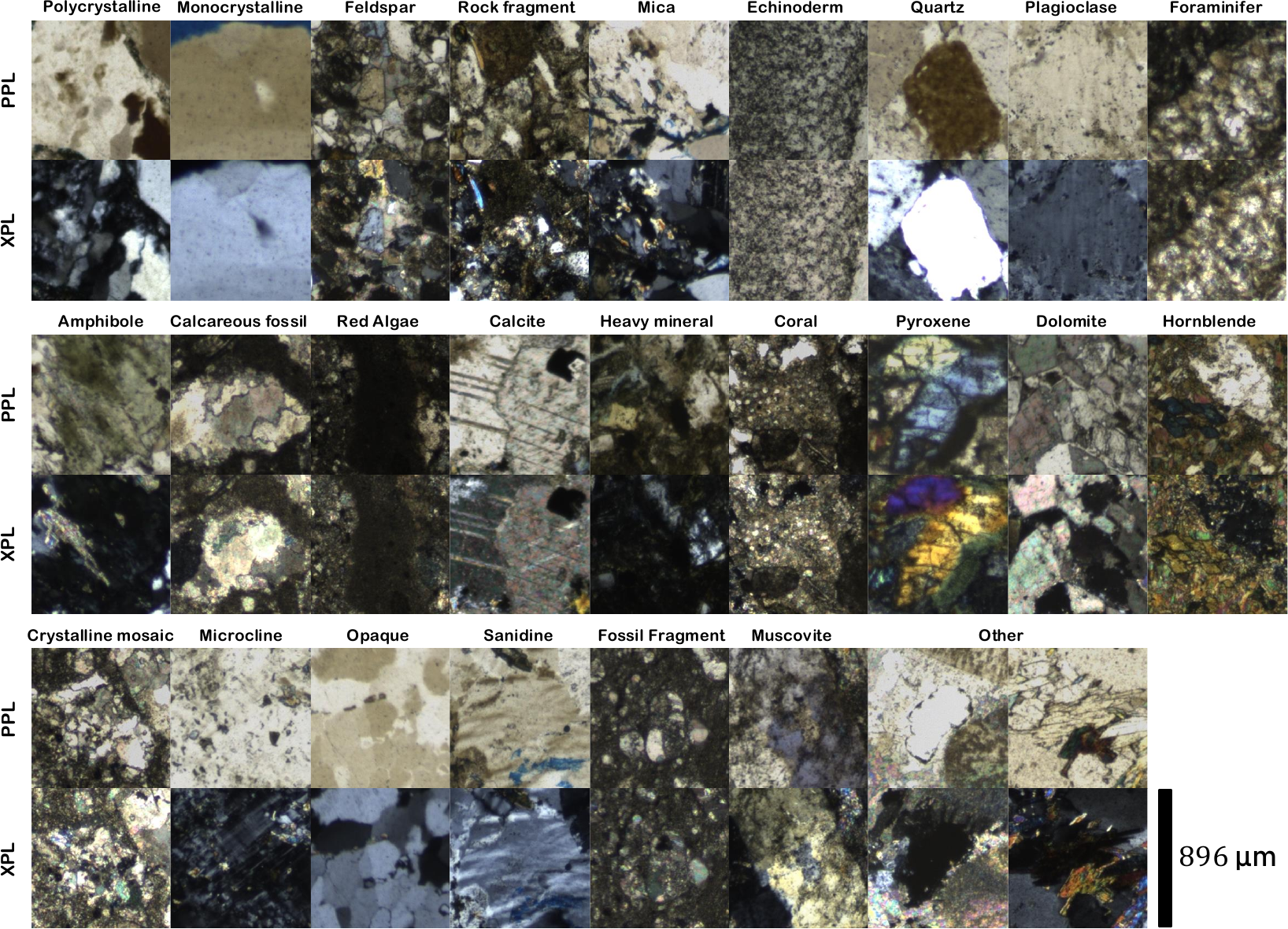}
    \caption{Example of 256 × 256 high-resolution image patches extracted from the LITHOS Dataset, illustrating the 25 mineral classes under the two polarization conditions: plane-polarized light (PPL) at 0° and cross-polarized light (XPL) at 0°. Each patch represents an area of $896 \mu m^2$. These paired images highlight the variation in color, texture, and birefringence patterns, which are critical for mineral identification in thin section petrography.}
    \label{fig:pols_smallIms}
\end{figure}

To assess a rock's mineralogical and textural characteristics, petrographers examine thin sections, which are thin flat slices of material prepared for microscopic analysis. Specifically, an approach known as the point-counting method is used. This technique superimposes a grid of equidistant points on the thin section. The petrographer examines each point, identifying and recording the mineral or feature present. This method allows for the determination of the modal composition of the rock, including the percentages of various minerals, porosity, and other constituents \citep{pointcounting}.

Despite its significance, petrographic analysis is a labor-intensive and expertise-driven task. With the growth of high-resolution imaging and modern computational tools, there is an opportunity to automate parts of this process, helping reduce subjectivity and improving efficiency. However, building reliable automated methods requires large, well-annotated datasets, which are currently lacking. Most existing databases are limited in size and variety, making it challenging to train robust models for this type of analysis \citep{L32,L51,L25,IL3}. Therefore, there is also a lack of publicly accessible methods that researchers can readily adopt and apply in this area.

This paper introduces the Large-scale Imaging and Thin section Optical-polarization Set (LITHOS), a novel experimental framework for automated petrography. LITHOS is the largest and most diverse publicly available resource for this task, surpassing existing frameworks by two orders of magnitude in the number of images and by one in the number of annotated mineral grains. Each mineral grain refers to an individual crystal or crystal fragment that can be visually distinguished in a thin section under a petrographic microscope based on its optical properties. The dataset encompasses 25 mineral categories, totaling 105,802 annotated grains and 211,604 high-resolution image patches. Each section was digitized using an automated petrographic microscope, capturing images under two polarization conditions. Expert annotations include mineral labels, grain size measurements, major and minor axes represented as intersecting vector paths, and spatial coordinates. Figure~\ref{fig:pols_smallIms} presents representative samples from the 25 categories under both polarization conditions. These examples illustrate the dataset's diversity in color, texture, and birefringence patterns. Additionally, they underscore the inherent complexity of the task, where several mineral categories exhibit significant visual similarities. This challenge reflects real-world petrographic analysis scenarios and makes LITHOS a robust benchmark for developing and evaluating automated mineral classification methods. This resource is intended to support the advancement of automated petrographic analysis, fostering further research in the field.

We train and evaluate a suite of standard deep learning models commonly used for image classification tasks on a single-polarization subset of LITHOS. Additionally, we introduce the LITHOS Baseline, a model specifically designed to leverage the dual-polarization imaging captured in our benchmark. The LITHOS Baseline is a transformer-based architecture that processes paired views of thin section patches under plane-polarized (PPL) and cross-polarized (XPL) light. This dual-encoder model exploits the complementary optical characteristics revealed by the two polarization modes, enabling more accurate mineral identification. 

Our main contributions can be summarized as follows:

\begin{enumerate}[(1),nosep,leftmargin=*,widest=*8]

    \item We collect the LITHOS Dataset, the largest publicly available image collection for automated petrography. We digitized thin sections under two polarization conditions, PPL and XPL, yielding 211,604 paired high-resolution image patches and 105,802 expert annotations across 25 mineral classes, representing the most comprehensive and diverse dataset of its kind. Expert annotations include mineral categories, grain coordinates, and geometric paths representing major and minor axes.
    \item We introduce the LITHOS Benchmark, a robust framework designed to evaluate automated mineral identification capabilities. The benchmark includes both binary and multi-class classification tasks. Additionally, we evaluate a set of deep learning single-polarization methods on this Benchmark.
    \item We propose the LITHOS Baseline, a dual-encoder transformer architecture specifically designed to leverage paired PPL and XPL views of mineral grains. Our baseline consistently improves mineral classification metrics over single-modality approaches, demonstrating the advantage of incorporating complementary polarization information in automated petrography.

\end{enumerate}


The LITHOS Dataset and pretrained models will be publicly released under a Creative Commons Attribution Non Commercial ShareAlike 4.0 International (CC BY-NC-SA 4.0) license upon acceptance, to promote reproducibility, transparency, and enable further research.

\section{Related Work}
\subsection{Traditional Petrography}

Petrographic analysis involves arduous and time-consuming procedures by trained personnel using specialized equipment \citep{pointcounting,thinsection_pro}. The process begins with the preparation of thin sections that are essential for microscopic examination. To obtain them, a rock sample is selected and cut into a small block using a diamond saw. This block is then sliced to obtain a thin sliver, typically around 1 mm thick. The fragment is ground flat and polished using progressively finer abrasives to achieve a smooth, optically flat surface. Subsequently, the polished sample is fixated to a glass slide with an adhesive, such as epoxy resin or Canada balsam. The mounted sample is ground to a standard thickness of approximately 30 µm. This process can take around 12 hours per thin section. Then, these thin sections are analyzed using a petrographic microscope equipped with polarizing filters. These filters produce polarized light that interacts with the properties of minerals, revealing critical optical characteristics. 

The petrographer examines each thin section under both PPL and XPL \citep{thinsection_pro}. By observing light intensity and color changes, the petrographer can identify minerals and assess their properties. For instance, a critical property observed during petrographic analysis is the extinction angle. This angle is the measure between a prominent crystallographic direction (such as cleavage planes or elongation) and the position at which the mineral goes dark (extinguishes) under XPL. Accurate measurement of the extinction angle provides insight into the mineral's internal structure and symmetry, which is essential for precise identification \citep{extAngle_LR1, extAngle_2}. The expert systematically repeats this process for numerous grains within each section using the point-counting method, which involves superimposing a grid of equidistant points onto the thin section. The petrographer examines each point, identifying and recording the mineral or feature present. The time required for analysis depends on the density of the grid and the petrographer's expertise. It is a common practice to analyze between 300 and 500 points per thin section to achieve a statistically valid representation. Reducing the number of points can accelerate the process, but may compromise the statistical robustness and accuracy of the results \citep{pointcounting}.

\subsection{Existing Machine Learning Benchmarks}

Recent advances in high-resolution imaging and computational modeling have driven interest in automating petrographic analysis. Despite these developments, progress remains constrained by the lack of large-scale, well-annotated datasets. Existing resources often fall short in capturing the mineralogical complexity required to support modern machine learning methods \citep{L32,L51,L25,IL3}. In particular, current datasets are often limited in the number of annotated samples, the diversity of mineral classes, the range of imaging modalities, or their public accessibility, all of which difficult training and evaluation of generalizable models.

\begin{table}[t]
\centering
\caption{Comparison between the LITHOS Dataset and existing petrographic datasets used for mineral identification. We report the polarizations available, the number of total images, the number of annotated grains, the number of mineral classes, and availability. A check mark (\checkmark) indicates presence, a dash (–) indicates absence, and (NR) indicates information Not Reported.}
\label{table:rw_comparison}
\resizebox{\textwidth}{!}{ 
\begin{tabular}{llllll}
\multicolumn{6}{l}{}                                      
\\ \hline 
Dataset                                       & Polarizations  & No. of Images & \begin{tabular}[c]{@{}c@{}}Annotated\\ Grains\end{tabular} &\begin{tabular}[c]{@{}c@{}}No. of \\ Classes\end{tabular} & \begin{tabular}[c]{@{}c@{}}Publicly\\ available\end{tabular}   \\ \hline \hline
{Hoque et al. \citep{L27}} & -             & 993       & NR                   & 4                               & -                  \\
\rowcolor[HTML]{EFEFEF} 
Song et al. \citep{L31-D}               & PPL, XPL           & 1,249     & 875                 & 10             & \checkmark                \\
Hongjue Li et al. \citep{L32}                   & \textbf{PPL, six XPLs}           & 140       & NR                   & NR          & -                  \\
\rowcolor[HTML]{EFEFEF} 
Keshk et al. \citep{L42}                       & PPL, XPL            & 1,388     & NR                   & 16                    & -                  \\
Ma et al. \citep{L50}                          & PPL, XPL        & 1,790     & 19,463              & 10                           & Upon request                \\
\rowcolor[HTML]{EFEFEF} 
Şener et al. \citep{L57}                        & XPL       & 600    & NR                   & 3                                & Upon request       \\ \hline \hline
LITHOS Dataset (Ours)                         & PPL, XPL          & \textbf{211,604}   & \textbf{105,802}             & \textbf{25}                       & \checkmark               
\end{tabular}
}
\end{table}

Several efforts have attempted to fill this gap, but important limitations persist. Table \ref{table:rw_comparison} shows existing petrographic datasets for
mineral identification. Datasets introduced by Hoque et al. \citep{L27}, Hongjue Li et al. \citep{L32}, and Keshk et al. \citep{L42} are not publicly available. Song et al. \citep{L31-D}, the only publicly available dataset, lacks either PPL or XPL images for some sections, limiting its utility for comprehensive model development. Şener et al. \citep{L57}  includes a very narrow set of mineral categories and images, limiting effective model training, the range of trainable architectures, and the ability to generalize to new mineral classes. Finally, although the proposal by Ma et al. \citep{L50} offers a valuable contribution to mineral semantic segmentation, the access to their full dataset requires direct request, limiting immediate usability for broader research and benchmarking. To overcome these drawbacks, we present the LITHOS Dataset: a publicly accessible resource featuring paired polarization images for all annotated grains, the broadest mineral class coverage to date, and a substantial volume of images and annotations suitable for training with state-of-the-art computer vision techniques.

As summarized in Table~\ref{table:rw_comparison}, LITHOS comprises 211,604 high-resolution RGB patches, and includes 105,802 mineral grain annotations spanning 25 classes. Compared to prior datasets, LITHOS provides two orders of magnitude more high-resolution images, one order of magnitude more labeled grains and broader class diversity. LITHOS includes the geometric paths of the major and minor axes for each grain. These axes provide a form of weak supervision that supports instance-level learning tasks. This representation offers a meaningful compromise between coarse annotations and full segmentation, enabling models to leverage spatial and structural representations without requiring exhaustive manual labeling. These characteristics make LITHOS the most comprehensive publicly available dataset for supervised learning in automated petrography.

Beyond dataset limitations, recent efforts have explored the use of deep learning for petrographic image analysis, particularly for tasks such as mineral classification and grain segmentation. Early studies employed conventional architectures such as CNNs \citep{L40, L54}, Faster RCNN \citep{L31}, U-Net \citep{L27, L29, L40, L42}, ResNet \citep{L26}, Mask2Former \citep{L57}, DeeplabV3+ \citep{L57} and Segformer \citep{L57} for segmentation and classification tasks, yet often lacked adaptation to the dual-polarization modality. Furthermore, the Multi-channel Attention Transformer proposed by \citep{L32} for rock segmentation, which takes into account multiple polarizations, lacks public code or pretrained models. Similarly, Trans-SedNet \citep{L50}, a dual-modal Vision Transformer, cannot be fully reproduced using the publicly released implementation. As a result, there remains a gap in the development of reproducible, polarization-aware deep learning models that are both open and extensible for broader applications in automated petrography. To address these limitations, we propose the LITHOS Baseline, an open source deep learning model adapted for dual-polarization processing and release the pretrained weights of our models so it can serve as a basis for multiple case studies in petrography.

\section{The Large-scale Imaging and Thin section Optical-polarization Set}
This section describes the construction of The Large-scale Imaging and Thin section Optical-polarization Set (LITHOS). First, we describe the thin sections collection, digitization and annotation process (Section \ref{dataset_collection}). Then, we define our benchmark for mineral classification and evaluation framework (Section \ref{classification_tasks}).

\subsection{LITHOS Dataset} \label{dataset_collection}

We compiled and annotated a total of 580 thin sections from Colombian soils, predominantly composed of formations and wells of interest for hydrocarbons exploration.

\begin{figure}[h!]
    \centering
    \includegraphics[width=0.7\linewidth]{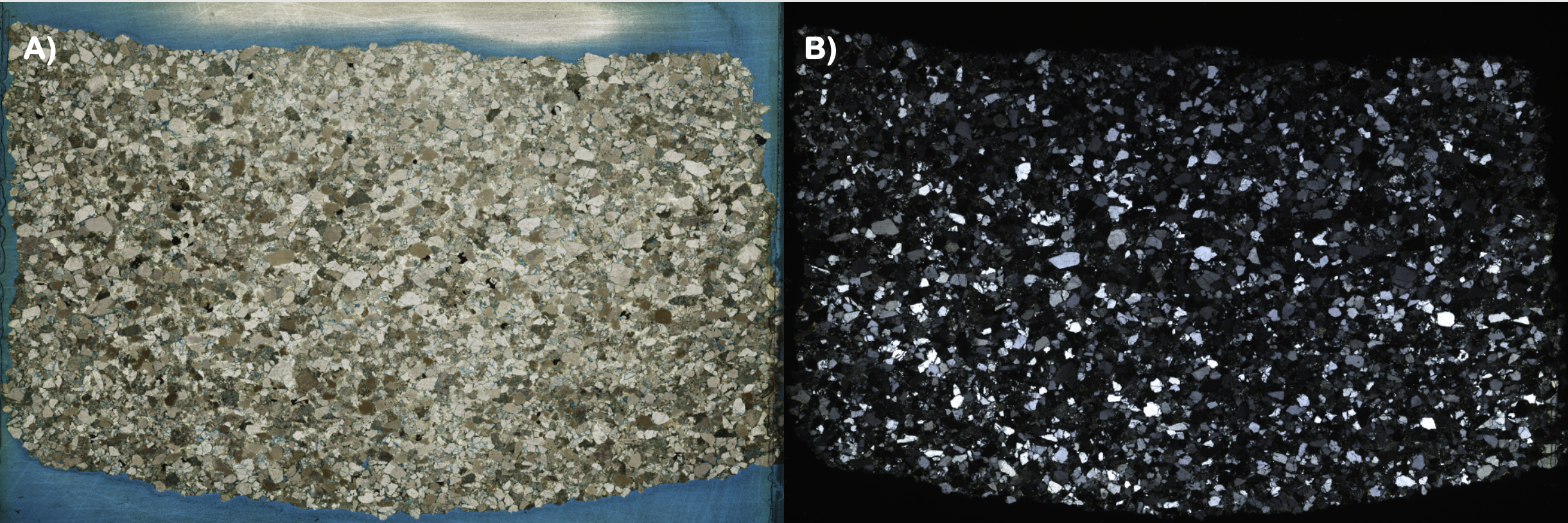}
    \caption{Example of a digitized thin section under polarized light. (A) Plane-polarized light (PPL) at 0°. (B) Cross-polarized light (XPL) at 0°.}
    \label{fig:pols}
\end{figure}

\textbf{Thin Sections Digitization.} We digitized each thin section using an automated petrographic microscope, which captures high-resolution images under different polarization angles \citep{OneGeo}. During imaging, the thin section is placed on the microscope stage, with a polarizer positioned below the sample and an analyzer above it. By rotating either the stage or the polarizing filters, the microscope can capture images in PPL (plane-polarized light)  and XPL (cross-polarized light). Specifically, images were acquired in PPL and XPL at 40X magnification. Each pixel represents an area of approximately $3.5 \mu m \times 3.5 \mu m$. For our dataset, we obtained two images per thin section: one PPL at 0° and one XPL at 0°, as shown in Fig.~\ref{fig:pols}. The image acquisition process involves an average of 143 acquisition points per thin section, with 20 images captured at each point, resulting in approximately 2,860 raw images per section. These images are processed using image alignment algorithms that automatically register and stitch them into coherent high-resolution mosaics (24,786 × 16,259 pixels) representing the entire thin sections. On average, the acquisition and initial processing requires 3 hours per sample. This pipeline, enables accurate reconstruction of large petrographic samples and supports downstream annotation on the OneGeo software \citep{OneGeo}. The dataset reflects a significant technological and human effort, corresponding to more than 1,700 hours of high-resolution image capture and processing.

\textbf{Annotation Process.} OneGeo \citep{OneGeo} was used to annotate all thin sections. This software functions as a petrographic microscope, utilizing the previously captured images to enable multi-magnification analysis. It also allows users to switch between polarization angles and visualize parallel and cross nicols simultaneously. These capabilities ensure a precise mineral identification. OneGeo generates an annotation grid of 300 to 600 equidistant points over each sample, and each point is individually analyzed. The annotation process for each 600-point thin section took around 10 hours. It included assigning a mineral category per grain, recording the x and y coordinates, measuring their minor and major axes, and representing its geometric paths as intersecting vectors through an HTML element. Each annotation undergoes a review by a second annotator, which typically requires approximately half the time of the initial annotation process. Detailed explanation of the annotation files can be found in the supplementary material (Listing 1). 

\textbf{Dataset}. In total, our database consists of 1,164 high-resolution RGB images, 105,802 annotated grains, and 25 mineral categories. This effort represents the equivalent of a full-time expert dedicating an entire year exclusively to annotating our database, highlighting the substantial value and scale of our contribution. The mineral categories exhibit a long-tail distribution dominated by quartz. The dataset is highly imbalanced, with $49,8\%$ of the annotated points representing quartz, categorized into three main classes: Quartz, Monocrystalline, and Polycrystalline. From each high-resolution image, we extracted patches of size 256 × 256 pixels. Each patch was centered at the intersection of the major and minor axes of the mineral. The axes paths were stored in an HTML element and reconstructed using the svgpathtools library in Python. This results in a total of 211,604 high-resolution RGB patches (Table \ref{table:rw_comparison}).

\subsection{LITHOS Benchmark} \label{classification_tasks}

\begin{center}
\begin{minipage}{0.45\linewidth}
\centering
\includegraphics[width=\linewidth]{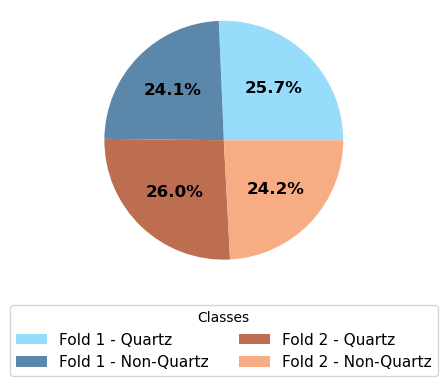}
\captionof{figure}{Distribution of annotated minerals across the two folds in the binary task.}
\label{fig:minerals_distr_bin}
\end{minipage}%
\hfill
\begin{minipage}{0.48\linewidth}
The LITHOS Benchmark is designed to assess mineral classification performance in soil thin section images. Considering the long-tail distribution of our dataset dominated by quartz, we propose two classification tasks. In the binary task, the goal is to classify the central mineral grain in each 256×256 image patch as either Quartz vs. Non-Quartz. The Quartz class aggregates several quartz varieties, including Monocrystalline, Polycrystalline, and generic Quartz annotations, while Non-Quartz aggregates the remaining classes. Fig \ref{fig:minerals_distr_bin} shows folds distribution for this setup. 

\vspace{\baselineskip}
In the multi-class task, the objective is to classify the central grain into one of 25 mineral categories. This ensures that the benchmark supports the development of more robust and generalizable AI models.
\end{minipage}
\end{center}

\begin{figure}[h!]
    \centering
    \includegraphics[width=1\linewidth]{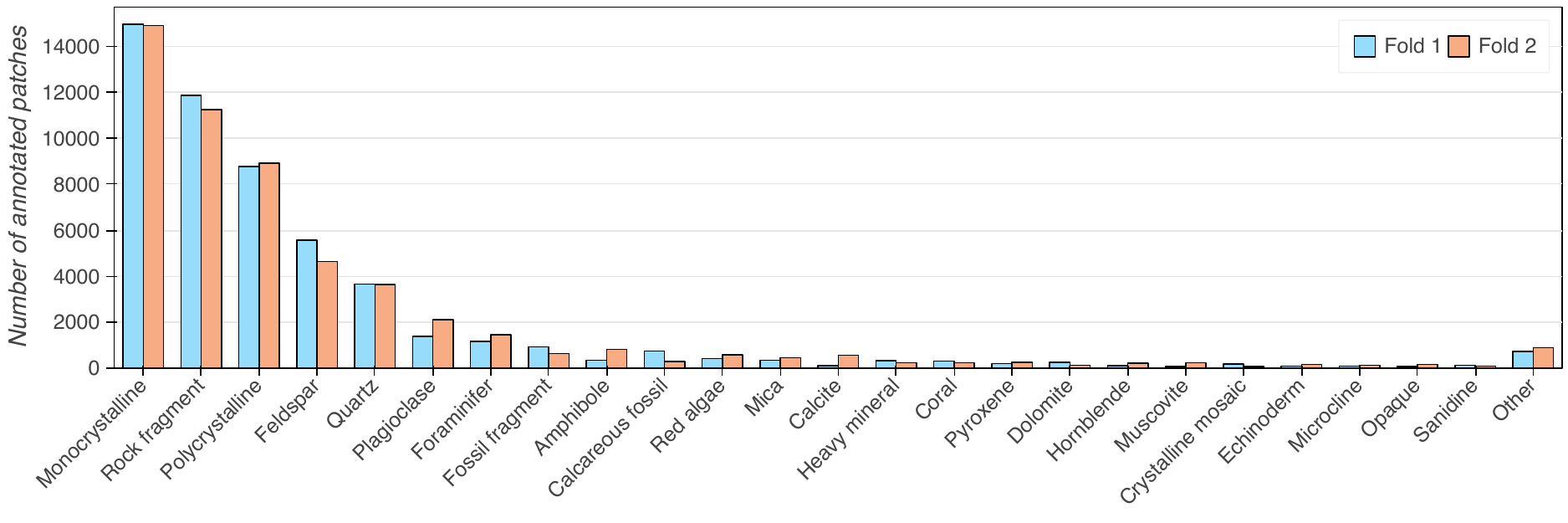}
    \caption{\textbf{Distribution of annotated minerals per class across the two folds.} The distribution is imbalanced, with a few dominant classes such as Monocrystalline, Rock Fragment, and Polycrystalline accounting for the majority of annotations. This long-tailed distribution reflects the natural occurrence of minerals in thin sections and poses a significant challenge for learning robust classification models, particularly for rare classes.}
    \label{fig:minerals_distr}
\end{figure}

We implemented a 2-fold cross-validation setup, assigning each thin section exclusively to one of the folds to avoid data leakage. This strategy preserved the relative mineral distribution across folds. The distribution of mineral categories across folds is shown in Fig. \ref{fig:minerals_distr}. Model performance of both tasks is evaluated using standard classification metrics: accuracy, precision, recall, and F1-score.

\section{LITHOS Baseline}
\label{lithos-baseline}

\begin{figure}[h!]
    \centering
    \includegraphics[width=0.80\linewidth]{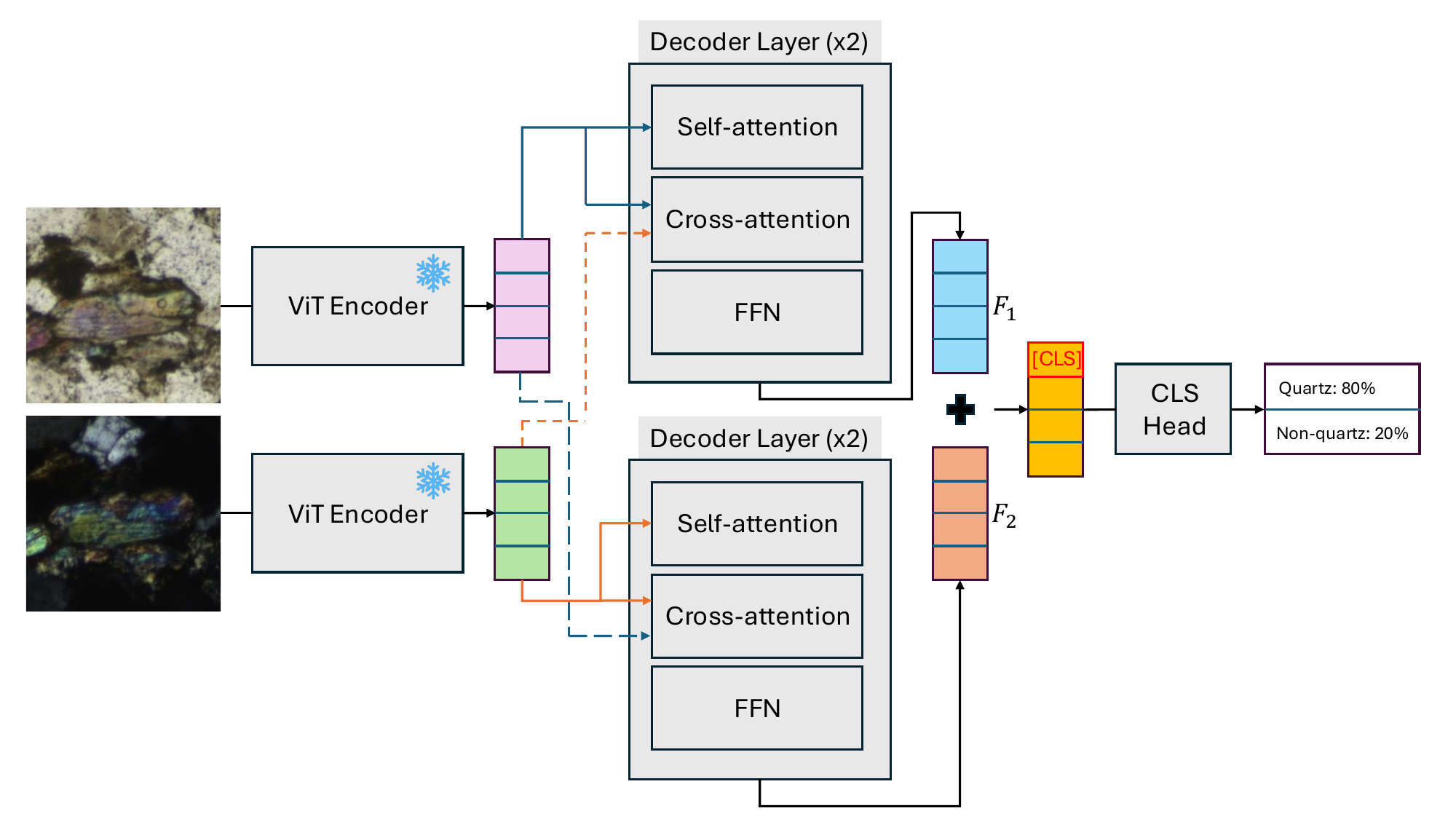}
    \caption{\textbf{Overview of the LITHOS Baseline.} Two frozen ViT encoders extract specific polarized representations from petrographic images at PPL and XPL. A dual-decoder module captures feature dependencies through self-attention and cross attention mechanisms. These features are then fused together via a learnable weighted sum. Lastly, the \textcolor{red}{[CLS]} token of the combined representation is passed through the classification head of the model. FFN stands for Feed-Forward Network.}
    \label{fig:model}
\end{figure}

For mineral classification in the LITHOS Benchmark, we propose the LITHOS Baseline (Fig. \ref{fig:model}), a transformer-based model designed to take advantage of multipolarization petrographic images. This model takes as input the PPL and XPL images of the same thin section patch to capture complementary mineralogical features.

Firstly, we pretrained two single-polarization ViTs \citep{dosovitskiy2020image} on the LITHOS dataset, each specializing on either PPL or XPL images. Then, we extracted their encoders and froze them to preserve their learned polarization-specific features. This design encourages each encoder to focus on distinct, complementary representations. The frozen pretrained encoders  independently transform each image into deep feature representations capturing mineral structures under distinct polarization conditions. Then, to combine the extracted representations, we introduce a dual-decoder architecture inspired by cross modality transformers \citep{tan2019lxmert, zheng2020cross}. 
Each decoder comprises two consecutive layers that apply self-attention within their assigned feature set and cross-attention to features from the other polarization, followed by a feed-forward network (FFN). 

We consider two different feature fusion strategies: (1) concatenating the outputs of both decoders along the feature dimension followed by a linear projection into a common space before classification, and (2) computing a weighted sum of the decoder outputs using a single learnable parameter $\alpha \in [0,1]$ such that the combined feature is $F_{\text{combined}} = \alpha  F_1 + (1-\alpha) F_2$, where $F_1$ and $F_2$ are the corresponding decoder outputs. We adopt the latter, $\alpha$-weighted summation as the fusion strategy, as it involves a single trainable parameter, directly combines the features, and maintains similar performance metrics compared to using the linear projection layer.

Finally, we pass the [CLS] token from $F_{\text{combined}}$ through a fully connected classification head, which predicts the mineral type based on the integrated multi-polarization features. We optimize the model with the Cross Entropy loss, commonly employed for classification tasks.

\textbf{Implementation details.} We trained all variants of the model on a single Quadro RTX 8000 GPU, equipped with 48 GB of VRAM. For all experiments, we used a batch size of 64 and trained each model for 10 epochs. In the binary classification task (Quartz vs. Non-Quartz), we used the Adam optimizer and StepLR scheduler, a learning rate of $1.3e-4$, momentum of $0.88$, and weight decay factor ($\gamma$) of $0.66$. In the multi-class setting, we used the SGD optimizer with CosineAnnealingLR scheduling. The initial learning rate was set to $8.9e-4$, momentum to $0.67$, and $\gamma$ to $0.89$ . The multi-class model was initialized from the pretrained weights of the binary classification task. Each single-polarization ViT experiment took around 10 hours, whereas the dual-polarization experiments took approximately 18. Our LITHOS baseline has around 674M parameters, from which only 10\% remain unfrozen during training.

\section{Results and Discussion}

Table \ref{table:binaryclass} presents the results of our binary mineral classification task (Quartz vs. Non-quartz), comparing the performance of several deep learning architectures over our 2-fold cross-validation setup. We trained single-polarization baselines architectures, including ResNet, GoogLeNet, ViT, and Swin Transformer using either PPL or XPL, with our proposed LITHOS Baseline incorporating both modalities. Among the single-modality baseline architectures, ViT achieved the highest recall ($0.894 \pm 0.021$), whereas Swin Transformer achieved the best F1-score ($0.855 \pm 0.006$) when trained with XPL. However, the LITHOS Baseline consistently outperformed the single-modality methods on multiple metrics, including the F1 score ($0.861 \pm 0.005$), precision ($0.835 \pm 0.013$), and accuracy ($0.851 \pm 0.002$).

\begin{table}[h!]
\centering
\caption{Binary classification results (Quartz vs. Non-Quartz). Models trained using either single or dual polarization modalities are evaluated using accuracy (Acc), recall, precision, and F1-score. Our LITHOS Baseline shows consistent improvements across various classification metrics compared to single-modality baselines. \textbf{Bold} indicates the best result, and {\ul underlined} indicates the second best.}
\label{table:binaryclass}
\resizebox{\textwidth}{!}{ 
\begin{tabular}{lllllll}
\hline
\multicolumn{7}{c}{\textbf{Binary Classification Task}}                                 \\ \hline
\multicolumn{1}{c}{\textbf{DL Architecture}} & \textbf{PPL} & \textbf{XPL} & \textbf{Acc}  & \textbf{Recall} & \textbf{Precision} &
\textbf{F1-score} \\
\hline
\textbf{Resnet \cite{he2016deep}}                                                      & \cmark          & \xmark           & $0.819 \pm 0.017$               & $0.882 \pm 0.037$          & $0.792 \pm 0.003$           & $0.834 \pm 0.018$              \\
\textbf{Googlenet \cite{szegedy2015going}}                                                   & \cmark          & \xmark           & $0.826 \pm 0.018$                  & $0.855 \pm 0.021$          & $0.817 \pm 0.014$           & $0.836 \pm 0.017$              \\
\textbf{ViT \cite{dosovitskiy2020image}}                                                         & \cmark          & \xmark           & $0.834 \pm 0.018$                  & {\ul $0.893 \pm 0.012$}          & $0.807 \pm 0.017$           & $0.848 \pm 0.016$              \\
\textbf{ViT \cite{dosovitskiy2020image}}                                                         & \xmark           & \cmark          & $0.840 \pm 0.002$                 & $\mathbf{0.894 \pm 0.021}$          & $0.815 \pm 0.013$  & $0.852 \pm 0.002$            \vspace{1pt} \\ 
\textbf{Swin Transformer\cite{liu2021swin}}                                                         & \cmark           & \xmark          & $0.835 \pm 0.014$                & $0.872 \pm 0.016$          & $0.821 \pm 0.011$  & $0.845 \pm 0.013$             \vspace{1pt} \\ 
\textbf{Swin Transformer\cite{liu2021swin}}                                                         & \xmark           & \cmark          & {\ul$0.843 \pm 0.008$}                & $0.892 \pm 0.001$          & {\ul$0.822 \pm 0.013$}  & {\ul$0.855 \pm 0.006$}             \vspace{1pt} \\ \hline
\textbf{LITHOS Baseline}                                   & \cmark          & \cmark          & $\mathbf{0.851 \pm 0.002}$    & $0.888 \pm 0.023$ & $\mathbf{0.835 \pm 0.013}$     & $\mathbf{0.861 \pm 0.005}$        \vspace{1pt} \\ \hline
\end{tabular}
} 
\end{table}

\begin{table}[h!]
\centering
\caption{Multi-class mineral classification results across 25 categories. 
The models are evaluated using accuracy (Acc), and macro-averaged recall, precision, and F1-score. 
Our LITHOS Baseline achieves the highest classification scores across all metrics. 
\textbf{Bold} indicates the best result, and {\ul underlined} indicates the second best.}

\label{table:multi}
\resizebox{\textwidth}{!}{ 
\begin{tabular}{lllllll}
\hline
\multicolumn{7}{c}{\textbf{Multi-Classification Task}}                                 \\ \hline
\multicolumn{1}{c}{\textbf{DL Architecture}} & \textbf{PPL} & \textbf{XPL} & \textbf{Acc}   & \textbf{Recall} & \textbf{Precision} &
\textbf{F1-score} \\
\hline
\textbf{Resnet \cite{he2016deep}}                                                      & \cmark          & \xmark           & $0.573 \pm 0.025$         & $0.342 \pm 0.024$          & $0.412 \pm 0.018$          & $0.343 \pm 0.013$                         \\
\textbf{Googlenet \cite{szegedy2015going}}                                                   & \cmark          & \xmark           & $0.579 \pm 0.029$          & $0.367 \pm 0.037$          & $0.459 \pm 0.009$          & $0.380 \pm 0.030$                      \\
\textbf{ViT \cite{dosovitskiy2020image}}                                                         & \cmark          & \xmark           & $0.573 \pm 0.019$          & $0.353 \pm 0.018$          & $0.436 \pm 0.032$          & $0.360 \pm 0.018$                    \\
\textbf{ViT \cite{dosovitskiy2020image}}                                                         & \xmark           & \cmark          &  $0.586 \pm 0.009$          & $0.365 \pm 0.030$          &  $0.459 \pm 0.008$         &  $0.382 \pm 0.025$             \vspace{1pt} \\ 
\textbf{Swin Transformer \cite{liu2021swin}}                                                         & \cmark           & \xmark          & $0.593 \pm 0.025$          &  $\mathbf{0.421 \pm 0.016}$         & $0.470 \pm 0.006$         & $0.422 \pm 0.010$             \vspace{1pt} \\ 
\textbf{Swin Transformer \cite{liu2021swin}}                                                         & \xmark           & \cmark          & {\ul $0.604 \pm 0.017$}          & $\mathbf{0.421 \pm 0.001}$          & {\ul $0.472 \pm 0.039$}         & {\ul$0.424 \pm 0.011$}             \vspace{1pt} \\ 
\hline
\textbf{LITHOS Baseline} & \cmark          & \cmark   & $\mathbf{0.623 \pm 0.019}$ & {\ul$0.415 \pm 0.006$} & $\mathbf{0.519 \pm 0.001}$ & $\mathbf{0.432 \pm 0.002}$       \vspace{1pt} \\ \hline
\end{tabular}
} 
\end{table}

Table \ref{table:multi} presents the results of our multi-class mineral classification task.
Similar to the binary classification setup, we conducted all experiments using a 2-fold cross-validation strategy, ensuring a reliable assessment of each model's generalization performance. Given the 25-class setting and the highly imbalanced nature of the dataset (as shown in Fig.~\ref{fig:minerals_distr}), we report macro-averaged precision, recall, and F1-score to provide an unbiased evaluation across all classes.
Among the single-polarization baselines, the Swin Transformer trained with XPL achieves the highest F1-score ($0.424 \pm 0.011$), recall ($0.421 \pm 0.001$), and precision ($0.472 \pm 0.039$). Likewise to the binary task, our LITHOS Baseline achieves the best overall performance across key metrics, including an accuracy of $0.623 \pm 0.019$, a precision of $0.519 \pm 0.001$, and an F1-score of $0.432 \pm 0.002$, demonstrating the benefits of polarization-based feature extraction.

These results highlight the effectiveness of leveraging both polarization modalities, with the LITHOS Baseline consistently outperforming single-modality models across both proposed classification tasks. Particularly noteworthy is the cross-attention mechanism integrated within our dual-decoder fusion module (Fig.~\ref{fig:model}), which enables effective interaction between the feature maps extracted from each polarization modality. This design facilitates richer representations and leads to more accurate classifications in both tasks.

Additional details on model complexity and inference performance are provided in the supplementary material (Table 1), where we show that the LITHOS Baseline, despite having a high total parameter count and number of FLOPs due to its dual ViT encoders, maintains a moderate training cost by keeping these encoders frozen. Precision-Recall curves for each class are also included in the supplementary material (Fig. 1 \& 2) for both classification tasks.

\begin{figure}
    \centering
    \includegraphics[width=0.8\linewidth]{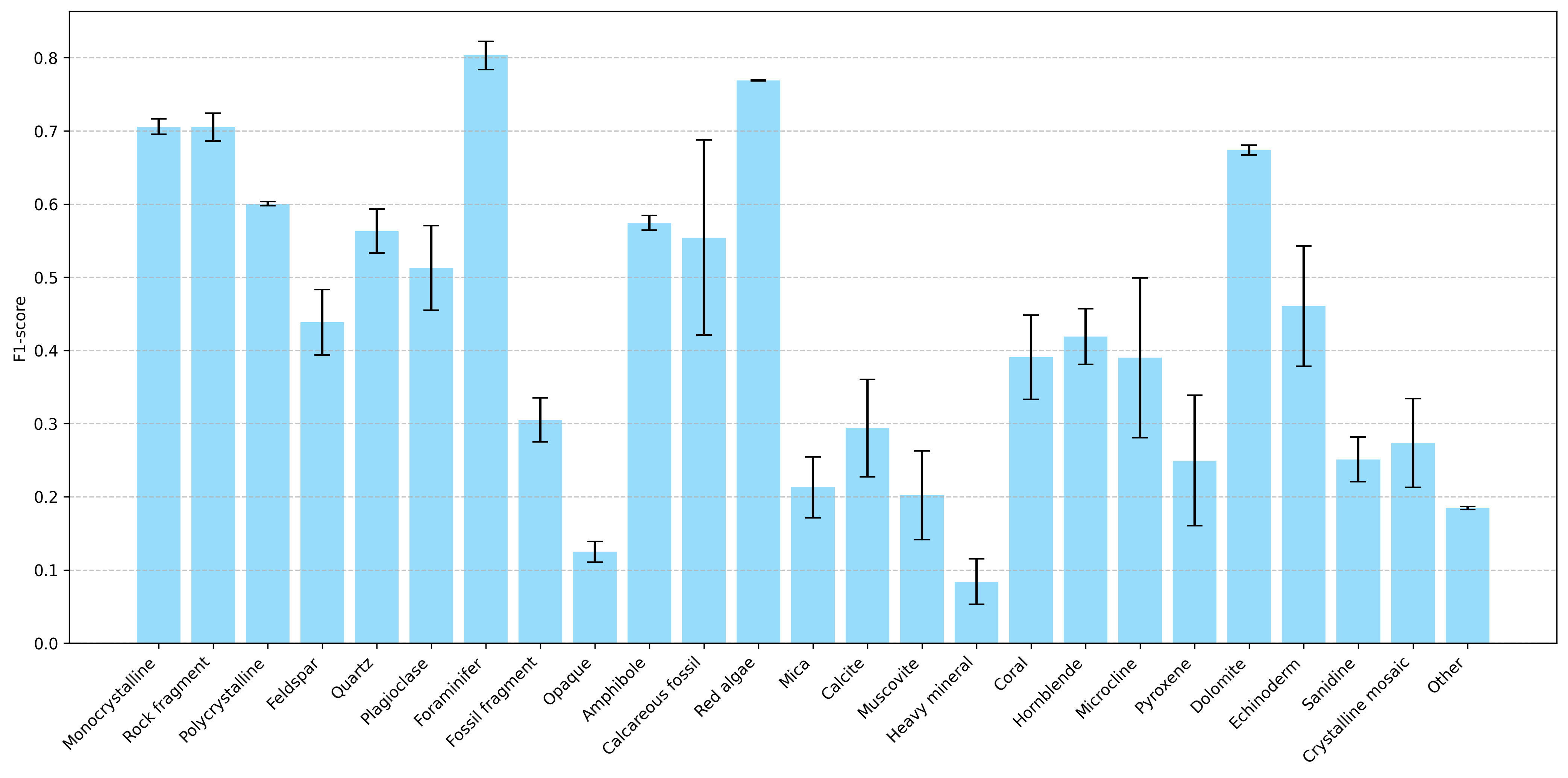}
    \caption{\textbf{Main Results of our proposed baseline model for the multi-classification task in the LITHOS Benchmark.} Bars represent the mean F1-score with standard deviation computed over the 2-folds for all 25 mineral categories.}
    \label{fig:f1multiclass}
\end{figure}

Even though we observe an overall performance improvement with our model, a more detailed per-class analysis (Fig. \ref{fig:f1multiclass}) reveals a significant long-tail challenge related to class imbalance. Classes such as Monocrystalline, Rock Fragment, and Polycrystalline dominate the dataset, often achieving F1-scores above $0.60$. However, low-frequency classes such as Opaque, Heavy mineral and Other, consistently exhibited F1-scores below $0.2$. This imbalance reflects the natural distribution of minerals in geological samples, but also limits model performance on underrepresented classes. 

We hypothesize that certain classes, such as Foraminifer and Red Algae, achieve notably high performance due to their distinctive biological structures, often characterized by porous textures or recognizable morphologies. Similarly, the model performs strongly on Dolomite, likely due to its unique rhombohedral shapes and cleavage or twinning patterns under XPL. In contrast, mid-performing classes often share overlapping visual characteristics under polarized light, especially in fine-grained or heterogeneous textures, making them harder to distinguish (eg. Feldspar, Quartz, and Plagioclase). Additionally, broader categories such as Heavy Mineral and Other encompass diverse and visually inconsistent instances, further contributing to classification challenges. 

Further insight into the model’s behavior is provided by the confusion matrices in Supplementary Fig. 3. A considerable proportion of the model’s misclassifications arise from confusions with the two most represented classes: Monocrystalline and Rock fragment. The Other class is misclassified across nearly all categories, likely due to its intrinsic heterogeneity. Regarding the confusions between Sanidine, Microcline, and Feldspar, it is important to note that Sanidine and Microcline are specific potassium feldspars, whereas the Feldspar class refers to feldspars in general. All three minerals are colorless, low-relief, and weakly birefringent under polarized light, and are primarily distinguished by their twinning patterns, which can be subtle at the patch level.

Similarly, the misclassification among Fossil fragment, Foraminifer, Coral, and Echinoderm is reasonable, as these biogenic carbonates share similar optical behavior under both PPL and XPL light. Although Dolomite exhibits strong performance overall, its confusion with Crystalline mosaic likely results from their comparable textural and optical features: both display interlocking, equigranular carbonate crystals forming a mosaic-like texture and appear colorless under PPL. Despite these limitations, our baseline demonstrates robustness and generalization across both folds, validating the value of polarization fusion and setting a new benchmark in automated petrographic classification.

\section{Limitations and Impact}
\label{limitations-impact}


Our work presents some limitations that offer avenues for future work. First, we acknowledge the geographical limitations of our dataset. However, Colombia exhibits a high degree of soil and geological diversity, which enhances the representativeness of our data to some extent. The country spans multiple climatic zones and varied topography, resulting in a broad range of soil types. Additionally, our collaborators collected these samples throughout the productive and potentially productive areas of the country. While this diversity does not fully compensate for the lack of data from other global regions, it provides a valuable starting point for broader applications. Additionally, the dataset exhibits a long-tail distribution, with nearly half of the annotations corresponding to quartz and its subtypes. This distribution reflects the natural abundance of these minerals in many geological contexts. We intentionally preserved this imbalance to ensure that models trained on the dataset are exposed to realistic mineral proportions encountered in practical applications, particularly in sustainable soil and rock exploration. While this long-tail distribution introduces challenges for multi-class classification, it also provides a more authentic benchmark for developing models capable of handling naturally imbalanced mineral compositions.

Regarding our proposed method, the supervision strategy of future methods could be improved by incorporating not only the intersection of the annotated grain-defining paths, but also the full extent of those paths, which represent the mineral’s minor and major axes. However, exploring such enhanced supervision was beyond this paper’s scope, as our primary goal was to establish a strong benchmark with baseline models based on single- and dual-modality polarization data using single-grain supervision. Additionally, the dataset exhibits class imbalance, a reflection of the natural distribution of minerals in soil, which suggests that further techniques could be applied to mitigate this and potentially improve performance metrics. Finally, our proposed LITHOS Baseline, is relatively large in terms of parameters because it combines two pretrained ViT encoders from single-modality baselines. Nevertheless, by freezing these encoders during training, we effectively reduce the computational burden.

The public release of LITHOS aims to democratize access to high-quality petrographic data, encourage reproducibility in mineral classification tasks, and stimulate interdisciplinary work between geoscientists and machine learning researchers. We believe that this work does not present ethical risks or negative societal impact and offers positive impact by enabling progress in automated petrography and geological analysis.

\section{Conclusion}
We introduced the Large-scale Imaging and Thin-section Optical-polarization Set (LITHOS) Benchmark, a comprehensive experimental framework that formalizes the problem of automated petrographic mineral classification by combining a large-scale, richly annotated dataset with a rigorous evaluation protocol and a set of baseline models. This benchmark defines a set of challenging binary and multi-class classification tasks grounded in real-world petrographic workflows, supported by expert annotations and dual-polarized imaging modalities. Our proposed LITHOS Baseline, which effectively integrates multi-modality polarization data, consistently outperformed single-modality baselines on both binary and multi-class setups, demonstrating the benefits of multi-view feature fusion. By publicly releasing the dataset, code, and pretrained models, we aim to establish a robust foundation that fosters reproducibility and stimulates interdisciplinary research towards automated petrography.

\section{Acknowledgements}
The authors gratefully acknowledge the financial support of the University of the Andes Foundation. This work was supported by Azure sponsorship credits granted by Microsoft´s AI for Good Research Lab. The authors further thank Cristhian Forigua, Natalia Valderrama, and Catalina Gómez Caballero for their valuable contributions in the initial stages of this project. Isai Daniel Chacón acknowledges the support of the 2023 UniAndes–DeepMind Scholarship.

\newpage

\bibliographystyle{plainnat}
\bibliography{Styles/bib}

\newpage
\section*{NeurIPS Paper Checklist}

\begin{enumerate}

\item {\bf Claims}
    \item[] Question: Do the main claims made in the abstract and introduction accurately reflect the paper's contributions and scope?
    \item[] Answer: \answerYes{}
    \item[] Justification: The main claims are justified because the abstract and introduction clearly present the LITHOS Benchmark as a comprehensive framework—including our novel public dataset, problem formulation, and baseline models (single- and double-modality). We explicitly state our key contributions at the end of the introduction, aligning fully with the paper’s scope.

\item {\bf Limitations}
    \item[] Question: Does the paper discuss the limitations of the work performed by the authors?
    \item[] Answer: \answerYes{}
    \item[] Justification: The paper discusses its limitations clearly in a dedicated "Limitations and Impact" section (Section \ref{limitations-impact}). This section acknowledges key assumptions and areas for future improvement, such as the scope of supervision strategies, the natural class imbalance in the dataset, and the computational demands of the proposed LITHOS baseline model. It also reflects on the practical implications of these factors and the trade-offs made, demonstrating transparency about the robustness and scope of the approach.

\item {\bf Theory assumptions and proofs}
    \item[] Question: For each theoretical result, does the paper provide the full set of assumptions and a complete (and correct) proof?
    \item[] Answer: \answerNA{} 
    \item[] Justification: There are no theoretical results, theorems, lemmas or proofs that should be stated or deduced in our paper.

    \item {\bf Experimental result reproducibility}
    \item[] Question: Does the paper fully disclose all the information needed to reproduce the main experimental results of the paper to the extent that it affects the main claims and/or conclusions of the paper (regardless of whether the code and data are provided or not)?
    \item[] Answer: \answerYes{}
    \item[] Justification: We provide public access to our code and data, so that people can reproduce our results. Regardless of this, we fully describe our proposed architecture in the LITHOS Baseline (Section \ref{lithos-baseline}) and show a diagram of it (Fig. \ref{fig:model}). In this section we also describe the resources we used and important implementation details for reproducibility purposes. Since a new dataset is also a contribution of our work, we fully describe the way in which we obtained our high resolution petrographic images and posterior RGB patches in the Section \ref{dataset_collection}.

\item {\bf Open access to data and code}
    \item[] Question: Does the paper provide open access to the data and code, with sufficient instructions to faithfully reproduce the main experimental results, as described in supplemental material?
    \item[] Answer: \answerYes{}
    \item[] Justification: We provide full open access to our data via Kaggle () and code in github with clear and descriptive READMEs that allow any person interested in our work and petrography to faithfully reproduce the main experimental results. 

\item {\bf Experimental setting/details}
    \item[] Question: Does the paper specify all the training and test details (e.g., data splits, hyperparameters, how they were chosen, type of optimizer, etc.) necessary to understand the results?
    \item[] Answer: \answerYes{}
    \item[] Justification: Section \ref{lithos-baseline} describes all the implementation details, hyperparameters and optimizer for training the LITHOS baseline. Also, our open access dataset in kaggle contains 2 csv folds regarding our 2-fold cross validation data split. In the Section \ref{classification_tasks} we also describe that we assigned each thin section exclusively to one of the folds to avoid data leakage. This allowed us to preserve the relative mineral distribution across folds.

\item {\bf Experiment statistical significance}
    \item[] Question: Does the paper report error bars suitably and correctly defined or other appropriate information about the statistical significance of the experiments?
    \item[] Answer: \answerYes{}
    \item[] Justification: We used a 2 fold-cross validation precisely for statistical significance and generalization assessment of our baseline models. All of the results we present in Table \ref{table:binaryclass} and Table \ref{table:multi} present the mean and standard deviation between the two folds. Also, our Figure \ref{fig:f1multiclass} shows the mean F1-Score and error bars for each mineral class in our dataset.

\item {\bf Experiments compute resources}
    \item[] Question: For each experiment, does the paper provide sufficient information on the computer resources (type of compute workers, memory, time of execution) needed to reproduce the experiments?
    \item[] Answer: \answerYes{}
    \item[] Justification: The paper provides information on computer resources in the Implementation Details paragraph from LITHOS baseline section \ref{lithos-baseline}.
    
\item {\bf Code of ethics}
    \item[] Question: Does the research conducted in the paper conform, in every respect, with the NeurIPS Code of Ethics \url{https://neurips.cc/public/EthicsGuidelines}?
    \item[] Answer: \answerYes{}
    \item[] Justification: : We reviewed the Neurips Code of Etichs and assure that the research conforms in every respect to it.

\item {\bf Broader impacts}
    \item[] Question: Does the paper discuss both potential positive societal impacts and negative societal impacts of the work performed?
    \item[] Answer: \answerYes{}
    \item[] Justification: The paper discusses the potential positive societal impacts of the work in the "Limitations and Impact" section (Section \ref{limitations-impact}), emphasizing how the public release of LITHOS aims to democratize access to high-quality petrographic data, promote reproducibility, and foster interdisciplinary collaboration between geoscientists and machine learning researchers. In this section we also describe that we do not foresee ethical risks or negative societal impacts associated with this research.
    
\item {\bf Safeguards}
    \item[] Question: Does the paper describe safeguards that have been put in place for responsible release of data or models that have a high risk for misuse (e.g., pretrained language models, image generators, or scraped datasets)?
    \item[] Answer: \answerNA{}
    \item[] Justification: The paper poses no such risks.

\item {\bf Licenses for existing assets}
    \item[] Question: Are the creators or original owners of assets (e.g., code, data, models), used in the paper, properly credited and are the license and terms of use explicitly mentioned and properly respected?
    \item[] Answer: \answerYes{}
    \item[] Justification: We credit all people and institutions involved in our data collection and investigation. We use CC BY-NC-SA 4.0 for out Dataset. We are the owners of both Dataset and Code being released.
    \item[] Guidelines:
    \begin{itemize}
        \item The answer NA means that the paper does not use existing assets.
        \item The authors should citep the original paper that produced the code package or dataset.
        \item The authors should state which version of the asset is used and, if possible, include a URL.
        \item The name of the license (e.g., CC-BY 4.0) should be included for each asset.
        \item For scraped data from a particular source (e.g., website), the copyright and terms of service of that source should be provided.
        \item If assets are released, the license, copyright information, and terms of use in the package should be provided. For popular datasets, \url{paperswithcode.com/datasets} has curated licenses for some datasets. Their licensing guide can help determine the license of a dataset.
        \item For existing datasets that are re-packaged, both the original license and the license of the derived asset (if it has changed) should be provided.
        \item If this information is not available online, the authors are encouraged to reach out to the asset's creators.
    \end{itemize}

\item {\bf New assets}
    \item[] Question: Are new assets introduced in the paper well documented and is the documentation provided alongside the assets?
    \item[] Answer: \answerYes{}
    \item[] Justification: All new assets introduced in the paper are thoroughly documented: the code is available on GitHub with detailed READMEs and inline comments, the dataset is hosted on Kaggle with complete metadata accessible on the webpage, and the annotation format is clearly explained in the supplementary material.

\item {\bf Crowdsourcing and research with human subjects}
    \item[] Question: For crowdsourcing experiments and research with human subjects, does the paper include the full text of instructions given to participants and screenshots, if applicable, as well as details about compensation (if any)? 
    \item[] Answer: \answerNA{}
    \item[] Justification: The paper does not involve crowdsourcing nor research with human subjects.

\item {\bf Institutional review board (IRB) approvals or equivalent for research with human subjects}
    \item[] Question: Does the paper describe potential risks incurred by study participants, whether such risks were disclosed to the subjects, and whether Institutional Review Board (IRB) approvals (or an equivalent approval/review based on the requirements of your country or institution) were obtained?
    \item[] Answer: \answerNA{}
    \item[] Justification: The paper does not involve crowdsourcing nor research with human subjects.

\item {\bf Declaration of LLM usage}
    \item[] Question: Does the paper describe the usage of LLMs if it is an important, original, or non-standard component of the core methods in this research? Note that if the LLM is used only for writing, editing, or formatting purposes and does not impact the core methodology, scientific rigorousness, or originality of the research, declaration is not required.
    \item[] Answer: \answerNA{} 
    \item[] Justification: LLM is used only for writing, editing, or formatting purposes and does not impact the core methodology, scientific rigorousness, or originality of the research.

\end{enumerate}

\end{document}


\maketitle


\section{Database organization}

The dataset introduced in the main paper is publicly available on Kaggle. The root directory contains two folders, $GMAS\_LOTE$ and $GMAS\_LOTE2$, which together include 580 thin-section samples (383 in the former, 197 in the latter). Each thin section is stored as a subfolder that contains multiple grain-level annotations. Each folder comprises three files at the grain level: PPL-0.png and XPL-0.png images captured under plane- and cross-polarized light, respectively, and an annotation file named $center\_component.csv$. This CSV file includes detailed attributes for each grain: $Multiclass\_label\_name$, $Multiclass\_label$ (numeric), $Binary\_label\_name$, $Binary\_label$ (Quartz vs. Non-Quartz), $Major-axis$ and $Minor-axis$ (grain size in mm), $x\_annot$, $y\_annot$ (coordinates of the annotated point based on the point-counting method), $x\_inter$, $y\_inter$ (intersection point of the annotated paths), and $path\_1$, $path\_2$ (HTML path elements). All coordinates and HTML paths are in reference to the original high-resolution mosaic image of the thin section. To convert these values to the 256×256 patch-level reference, one must center the coordinate system on (128, 128), corresponding to the transformed position of ($x\_inter$, $y\_inter$). The main folder also contains two partition files, $Fold1\_complete\_info\_allclasses.csv$ and $Fold2\_complete\_info\_allclasses.csv$, specifying the relative paths to the PPL-0.png image of each grain assigned to the respective fold. As described in the main text, all grains from a given thin section belong exclusively to either Fold 1 or 2. The code provided below illustrates the conversion of path HTML elements to vectors in the patch space.

\begin{lstlisting}[language=Python, caption=Convert HMTL paths to vectors.]
from svgpathtools import parse_path
import pandas as pd

    annot_info = pd.read_csv(annot_file)
    path_1 = annot_info["path_1"][0]
    path_2 = annot_info["path_2"][0]
    data_path = {'path_1':path_1,'path_2':path_2}
    
    center_coords_or = [annot_info["x_inter"],annot_info["y_inter"]]
    center_coord_patch = [patch_size/2, patch_size/2]
    
    dx = center_coord_patch[0] - center_coords_or[0]
    dy = center_coord_patch[1] - center_coords_or[1]
    
    line_points = {'path_1':[],'path_2':[]}
    both_paths = ['path_1','path_2']
    
    image = Image.open(image_path)
    draw = ImageDraw.Draw(image)
    
    for path_name in both_paths:
        path_x = data_path[path_name]
        path_x = parse_path(path_x)
        
        end_point = ((path_x.start.real*24786)+dx,
                        (path_x.start.imag*24786)+dy)

                        
        start_point = ((path_x.end.real*24786)+dx,
                        (path_x.end.imag*24786)+dy)
                        
        line_points[path_name] = [start_point,end_point]
        
    draw.line(line_points['path_1'], fill=(255,0,0), width=10)
    draw.line(line_points['path_2'], fill=(255,0,0), width=10)
\end{lstlisting}

\section{Architecture Complexity and Inference Performance}

\begin{table}[h!]
\centering
\caption{Model complexity and average inference time for binary (2-class) and multi-class (25-class) classification. Params in millions (M), FLOPs in billions (B).}
\label{table:modelcomplexity_compact}
\resizebox{\textwidth}{!}{%
\begin{tabular}{l c c c c}
\hline
\multicolumn{5}{c}{\textbf{Model complexity and inference performance}} \\ \hline
\textbf{DL Architecture} & \textbf{Task} & \textbf{Params (Trainable / Total)} & \textbf{FLOPs (B)} & \textbf{Avg. Time (ms)} \\
\hline
ResNet            & Binary      & 23M / 23M    & 4.14  & $4.46$  \\
GoogLeNet    & Binary      & 5M / 5M      & 1.51  & $5.18$  \\
ViT       & Binary      & 303M / 303M  & 59.7  & $17.45$ \\
Swin Transformer   & Binary      & 86.9M / 86.9M& 17.1  & $25.91$ \\
LITHOS Baseline                      & Binary      & 67M / 673M   & 133.26& 40.18 \\ \hline
ResNet             & Multi-class & 23M / 23M    & 4.14  & $4.46$  \\
GoogLeNet     & Multi-class & 5M / 5M      & 1.51  & $5.18$  \\
ViT      & Multi-class & 303M / 303M  & 59.7  & $17.60$ \\
Swin Transformer   & Multi-class & 86.9M / 86.9M& 17.1  & $26.07$ \\
LITHOS Baseline                      & Multi-class & 67M / 673M   & 133.28& 40.18 \\ \hline
\end{tabular}%
}
\end{table}

Table~\ref{table:modelcomplexity_compact} shows the model complexity and inference performance for all architectures evaluated in both binary and multi-class classification tasks. The table compares the number of trainable and total parameters, the computational cost in FLOPs, and the average inference time per image. The LITHOS Baseline has a large total parameter count and a high number of FLOPs compared to the other baselines, due to its dual ViT encoders. However, these encoders are kept frozen during training, meaning only about 10\% of the parameters are updated. This design reduces the effective training cost and memory consumption while leveraging strong, pretrained features from each polarization stream, which is key to improving classification performance.

\newpage
\section{Binary Precision-Recall Curves}
\label{gen_inst}

\begin{figure}[h!]
    \centering
    \includegraphics[width=\linewidth]{Figures/precision_recall_curve_mean_std+swin.png}
    \caption{Precision-Recall curves across both folds (mean $\pm$ std) for the binary Quartz vs. No-Quartz classification task. Our LITHOS Baseline achive the best performance in both folds.}
    \label{fig:pols}
\end{figure}

\section{Multiclass Precision-Recall Curves}

\begin{figure}[tbp]
    \centering
    \includegraphics[width=\linewidth]{Figures/PR_Curves_1_2.png}
    \label{fig:pols}
\end{figure}

\begin{figure}[]
    \centering
    \includegraphics[width=\linewidth]{Figures/PR_Curves_2_2.png}
    \caption{Precision-Recall curves across both folds (mean $\pm$ std) for the multiclass classification task for the 25 minerals in the LITHOS Dataset. Our LITHOS baseline achieves higher performance in the most represented mineral classes of our database. In some classes, there is a noticeable increase in performance on models that incorporate XPL images of the minerals, instead of PPL-only models. This behavior might be related to distinctive features that are revealed under cross-polarized light. For example, cross-hatched twinning patterns reveal microcline’s gridiron structure hidden in featureless PPL views.}
    \label{fig:pols}
\end{figure}

\section{Multiclass Confusion Matrix}

\begin{figure}[H]
    \centering
    \begin{subfigure}[b]{\textwidth}
        \centering        \includegraphics[width=\textwidth]{Figures/confmat_PolarViT_fold1.pdf}
        \label{fig:img1}
    \end{subfigure}


    \begin{subfigure}[b]{\textwidth}
        \centering
        \includegraphics[width=\textwidth]{Figures/confmat_PolarViT_fold2.pdf}
        \label{fig:img2}
    \end{subfigure}
    \caption{Confusion matrices of the LITHOS Baseline model for the 25-class mineral classification task across both data folds.}
    \label{one_below_other}
\end{figure}